\pdfoutput=1
\def\year{2021}\relax
\documentclass[letterpaper]{article} 
\usepackage{aaai21}  
\usepackage{times}  
\usepackage{helvet} 
\usepackage{courier}  
\usepackage[hyphens]{url}  
\usepackage{graphicx} 
\usepackage{natbib}  
\usepackage{caption} 
\usepackage{graphicx}
\usepackage{booktabs}
\usepackage{amsmath}
\usepackage{amsfonts}
\usepackage{multirow}

\usepackage{footmisc}
\usepackage{bm}
\usepackage{xspace}

\def\rz{{\textnormal{z}}}

\def\rve{{\mathbf{e}}}

\def\rvx{{\mathbf{x}}}

\def\rvz{{\mathbf{z}}}

\def\vmu{{\bm{\mu}}}

\def\vb{{\bm{b}}}
\def\vc{{\bm{c}}}

\def\ve{{\bm{e}}}
\def\vf{{\bm{f}}}

\def\vh{{\bm{h}}}

\def\vx{{\bm{x}}}

\def\vz{{\bm{z}}}

\def\mE{{\bm{E}}}

\def\mI{{\bm{I}}}

\def\mW{{\bm{W}}}

\def\mSigma{{\bm{\Sigma}}}

\DeclareMathAlphabet{\mathsfit}{\encodingdefault}{\sfdefault}{m}{sl}
\SetMathAlphabet{\mathsfit}{bold}{\encodingdefault}{\sfdefault}{bx}{n}

\def\sR{{\mathbb{R}}}

\def\sV{{\mathbb{V}}}

\newcommand{\E}{\mathbb{E}}

\newcommand{\element}{\ensuremath{x}\xspace}
\newcommand{\velement}{\ensuremath{\vx}\xspace}

\newcommand{\elementbefore}{\ensuremath{\element_-}\xspace}
\newcommand{\elementafter}{\ensuremath{\element_+}\xspace}

\newcommand{\nelementbefore}{\ensuremath{\tilde{\element}_-}\xspace}
\newcommand{\nelementafter}{\ensuremath{\tilde{\element}_+}\xspace}

\newcommand{\velementbefore}{\ensuremath{\velement_-}\xspace}
\newcommand{\velementafter}{\ensuremath{\velement_+}\xspace}

\newcommand{\changeReprFunc}{\ensuremath{f_\Delta}\xspace}
\newcommand{\editrepr}[2]{\ensuremath{\changeReprFunc(#1,#2)}\xspace}

\newcommand{\relement}{\ensuremath{\rvx}\xspace}
\newcommand{\relementbefore}{\ensuremath{\relement_-}\xspace}
\newcommand{\relementafter}{\ensuremath{\relement_+}\xspace}

\makeatletter
\DeclareRobustCommand{\cev}[1]{%
	\mathpalette\do@cev{#1}%
}
\newcommand{\do@cev}[2]{%
	\fix@cev{#1}{+}%
	\reflectbox{$\m@th#1\vec{\reflectbox{$\fix@cev{#1}{-}\m@th#1#2\fix@cev{#1}{+}$}}$}%
	\fix@cev{#1}{-}%
}
\newcommand{\fix@cev}[2]{%
	\ifx#1\displaystyle
	\mkern#23mu
	\else
	\ifx#1\textstyle
	\mkern#23mu
	\else
	\ifx#1\scriptstyle
	\mkern#22mu
	\else
	\mkern#22mu
	\fi
	\fi
	\fi
}

\title{Variational Inference for Learning Representations \\ of Natural Language Edits}

\author{
    Edison Marrese-Taylor, 
    Machel Reid,
    Yutaka Matsuo
    \\
}

\affiliations{
    Graduate School of Engineering\\
    The University of Tokyo\\
    \{emarrese,machelreid,matsuo\}@weblab.t.u-tokyo.ac.jp
}

\begin{document}

\maketitle

\begin{abstract}
    Document editing has become a pervasive component of the production of information, with version control systems enabling edits to be efficiently stored and applied. In light of this, the task of learning distributed representations of edits has been recently proposed.  With this in mind, we propose a novel approach that employs variational inference to learn a continuous latent space of vector representations to capture the underlying semantic information with regard to the document editing process. We achieve this by introducing a latent variable to explicitly model the aforementioned features. This latent variable is then combined with a document representation to guide the generation of an edited version of this document. Additionally, to facilitate standardized automatic evaluation of edit representations, which has heavily relied on direct human input thus far, we also propose a suite of downstream tasks, PEER, specifically designed to measure the quality of edit representations in the context of natural language processing.
\end{abstract}

\section{Introduction}

Editing documents has become a pervasive component of many human activities \cite{miltnerFlySynthesisEdit2019}. This is, to some extent, explained by the advent of the electronic storage of documents, which has greatly increased the ease with which we can edit them.


From source code to text files, specially over an extended period of time, users often perform edits that reflect a similar underlying change. For example, software programmers often have to deal with the task of performing repetitive code edits to add new features, refactor, and fix bugs during software development. On the other hand, right before a conference deadline technical papers worldwide are finalized and polished, often involving common fixes for grammar, clarity, and style \cite{yin_learning_2019}. In light of this, it is reasonable to wonder if it would be possible to automatically extract rules from these common edits. This has lead researchers to recently propose the task of learning distributed representations of edits \cite{yin_learning_2019}.

In this paper, we explore the performance of latent models in capturing properties of edits. Concretely, we introduce a continuous latent variable to model features of the editing process, extending previous work and effectively proposing a new technique to obtain representations that can capture holistic semantic information in the document editing process. Since inference in latent variable models can often be difficult or intractable, our proposal follows previous work framing the inference problem as optimization \cite{kingmaAutoEncodingVariationalBayes2014,bowmanGeneratingSentencesContinuous2016}, which makes it an Edit Variational Encoder (\textsc{EVE}). Since it is a known fact that latent variable models for text face additional challenges due to the discrete nature of language \cite{bowmanGeneratingSentencesContinuous2016}, in this paper, we also propose a specific mechanism to mitigate this issue.

In addition to proposing \textsc{EVE}, we also note that the empirical evaluation of edit representation has, so far, mainly been based on semi-automatic techniques. For example, including visual inspection of edit clusters or human evaluation of certain quality aspects of the representations. As these evaluations mechanisms are generally time consuming and labor intensive, in this paper we propose a set of extrinsic downstream tasks specifically designed to more comprehensively evaluate the quality of edit representations. Our motivation is to help advance research in this task by introducing a fully automatic, well-defined way to measure what the learned latent space is capable of capturing. Similar endeavors have been a key element in tracking progress and developing new approaches in computer vision \cite{russakovsky_imagenet_2015, antol_vqa_2015} and natural language processing \cite{wangGLUEMultiTaskBenchmark2018}. We draw inspiration from several relevant problems from the latter, and leverage resourced from three different tasks, namely Wikipedia editing, machine translation post-editing and grammatical error correction to present our evaluation scheme.

Our results indicate that evaluation metrics that are related to the task used to obtain edit representations are generally good predictors for the performance of these representations in downstream tasks, although not always. Compared to existing approaches, our model obtains better scores on the intrinsic evaluation, and the representations obtained by our approach can also consistently deliver better performance in our set of introduced downstream tasks. Our code and data are available on GitHub\footnote{\url{https://github.com/epochx/PEER}}.

\section{Related Work}

Learning distributed representations for edits was perhaps first proposed indirectly by \citet{loyola2017,jiang_towards_2017,jiang_automatically_2017}. These works note that source code changes, or commits, are usually accompanied by short descriptions that clarify their purpose \cite{Guzman:2014:SAC:2597073.2597118} and explore whether this information could be used to create a mapping between the commits and their descriptive messages. Models that further improve the performance in this task have also been proposed in the last few years. For example, \citet{loyola_content_2018} proposed ways to provide additional context to the encoder or constrain the decoder with mild success, and \citet{liu_generating_2019} augmented the sequence-to-sequence methods with a copy mechanism based on a pointer net obtaining better performance. \citet{liuNeuralmachinetranslationbasedCommitMessage2018} tackled the problem using an approach purely based on machine translation.

Recently, \citet{yin_learning_2019} have directly proposed to learn edit representations by means of a task specifically designed for those purposes. While their ideas were tested on both source code and natural language edits, the work of \citet{zhao_neural_2019} proposed a similar approach that is specifically tailored at source code with relatively less success.

In natural language processing, edits have been studied mainly in two contexts. On one hand, edits are useful for the problem of machine translation post-editing, where humans amend machine-generated translations to achieve a better final product. This task has been crucial to ensure that production-level machine translation systems meet a given level of quality \cite{speciaTranslationQualityProductivity2017}. Although research on this task has focused mainly on learning to automatically perform post-editing, some recent work has more directly addressed the problem of modelling different editing agents \cite{goisTranslator2VecUnderstandingRepresenting2019}
in an effort to understand the nature of the human post-editing process, which is key to achieve the best trade-offs in translation efficiency and quality.

On the other hand, edits have also been relevant in the context of English grammatical error correction (GEC). In this task, given an English essay written by a learner of English as a second language, the goal is to detect and correct grammatical errors of all error types present in the essay and return the corrected essay. This task has attracted recent interest from the research community with several shared tasks being organized in the last years \cite{ ngCoNLL2014SharedTask2014,bryantBEA2019SharedTask2019}.

Additionally, given the importance that edits play in crowd-sourced resources such as Wikipedia, there has also been work on indirectly learning edit representations that are useful to predict changes in the quality of articles, which is cast as an edit-level classification problem \cite{sarkarStRESelfAttentive2019}. Similarly, \citet{marrese-taylor_edit-centric_2019} proposed to improve quality assessment of Wikipedia articles by introducing a model that jointly predicts the quality of a given Wikipedia edit and generates a description of it in natural language.

In terms of the proposed model, our approach is related to autoencoders \citep{Rumelhart:1986we}, which aim to learn a compact representation of input data by way of reconstruction. Our approach is also related to variational autoencoders \citep{kingmaAutoEncodingVariationalBayes2014}, which can be seen as a regularized version of autoencoders, specifically \cite{bowmanGeneratingSentencesContinuous2016}, who introduced an RNN-based VAE that incorporates distributed latent representations of entire sentences. Our architecture is also similar to that of \citet{guptaDeepGenerativeFramework2018} who condition both the encoder and decoder sides of a VAE on an input sentence to learn a model suitable for paraphrase generation, but we depart from this classic VAE definition as our generative process includes two observable variables.

Finally, our proposals are also related to \citet{guuGeneratingSentencesEditing2018},
who proposed a generative model for sentences that first samples a prototype sentence from the training corpus and then edits it into a new sentence, with the assumption that sentences in a single large corpus can be represented as minor transformations of other sentences. Instead, in our setting, edits are clearly identified by two distinct versions of each item (i.e. $x_-$ and $x_+$), which we can regard as a parallel corpus. Although this approach also captures the idea of edits using a latent variable, doing so is not the main goal of the model. Instead, our end goal is precisely to learn a function that maps an edit (represented by the aforementioned $x_-$ and $x_+$) to a learned edit embedding space.

\section{Proposed Approach}

The task of learning edit representations assumes the existence of a set $x^{(i)} = \{ \elementbefore^{(i)} ,\elementafter^{(i)} \}$, where $\elementbefore^{(i)}$ is the original version of an object and $\elementafter^{(i)}$ is its form after a change has been applied. To model the applied change, i.e. the edit, we propose the following generative process:
{\fontsize{9.5}{10}\begin{equation}
    p(\relementafter|\relementbefore)
    = \int_{\rz} p(\relementafter, z | \relementbefore)d_{z} = \int_{\rz} p(\relementafter | z, \relementbefore) p(z)d_{z}
\end{equation}
}
In the above equation, \relementafter and \relementbefore are observed random variables associated to $\elementafter^{(i)}$ and $\elementbefore^{(i)}$ respectively, and $z$ represents our continuous latent variable. Since the incorporation of this variable into the above probabilistic model makes the posterior inference intractable, we use variational inference to approximate it. The variational lower bound for our generative model can be formulated as follows:
\begin{align}
    \text{ELBO}(\relementafter, \relementbefore)
     & = -\text{KL}\left[q(z)||p(z) \right] \nonumber                             \\
     & \quad + \E_{q(\rz)} \left[\log p(\relementafter|z,\relementbefore) \right]
    \label{eq:elbo}
\end{align}
In Equation \ref{eq:elbo}, $p(z)$ is the prior distribution and $q(z)$ is the introduced variational approximation to the intractable posterior $p(z | \relementbefore, \relementafter)$. We assume that the edits in our dataset are i.i.d., allowing us to compute the joint likelihood of the data as the product of the likelihood for each example. This assumption enables us to write the following expression:
\begin{equation}
    \log{p(x^{(i)}, \ldots, x^{(N)})} = \sum_{i=1}^{N}{\log{p(\elementafter^{(i)}|\elementbefore^{(i)}})}
\end{equation}
Finally, we can write \cite{zhangVariationalNeuralMachine2016}:
\begin{align}
    \log{p(\elementafter^{(i)}|\elementbefore^{(i)}})
     & \ge \text{ELBO}(\elementafter^{(i)}, \elementbefore^{(i)})                                                    \\
     & \ge \mathbb{E}_{ \rz\sim q(z)}\left[\log{p(\elementafter^{(i)} |\elementbefore^{(i)}, \rz)} \right] \nonumber \\
     & \quad - \text{KL}\left[q(\rz)\|p(\rz)\right]
\end{align}
From now on, we refer to $\elementbefore^{(i)}$ and $\elementafter^{(i)}$ as
$\elementbefore$ and $\elementafter$ respectively. We set $\velementbefore, \velementafter \in \sR^{d}$ as continuous vectors to be the representation of the original version of an element $\elementbefore$ and its edited form $\elementafter$, and $\rvz \in \sR^{d}$ to be a continuous random vector capturing latent semantic properties of edits. Our goal is to learn a representation function \changeReprFunc that maps an edit $ (\elementbefore, \elementafter)$ to a real-valued edit representation $\editrepr{\velementbefore}{\velementafter} \in \sR^n$. Following previous work, we utilize neural networks to estimate the following components of our generative process:
\begin{itemize}
    \item $q(\rz) \approx q_\phi(\rvz | \velementbefore, \velementafter)$ is our variational approximator for the intractable posterior, where $q_\phi$ denotes the function approximated by this neural network parameterized by $\phi$.
    \item $p(\velementafter | \elementbefore, \vz) \approx p_\theta(\elementafter | \velementbefore, \rvz)$, where $p_\theta$ denotes the function defined by the neural net and its dependence on parameters $\theta$.
\end{itemize}

Our model is optimized with the following loss function:
\begin{align}
    \mathcal{L}(\phi, \theta)
     & = \mathbb{E}_{ \rz\sim q_\phi}\left[\log{p_\theta(\elementafter | \velementbefore, \rvz)} \right] \nonumber \\
     & \quad- \text{KL}\left[ q_\phi(\rvz | \velementbefore, \velementafter) \| p(\rz) \right]
\end{align}

From our component definitions, it follows that the neural network parameterizing $p_\theta(\elementafter| \elementbefore, \vz)$ acts as a \textbf{variational neural editor} and is trained to minimize the negative log-likelihood of reconstructing the edited version of each element. On the other hand, the neural net that parameterizes the approximate posterior $q_\phi(\rz)$ minimizes the Kullback-Leibler divergence with respect to the prior $p(\rz)$. Since we have made this function depend explicitly on each edit, this component can be considered as a \textbf{variational neural edit encoder}.

Our loss function contains an expectation term computed over the random latent variable introduced. To be able to train our neural components using backpropagation, we utilize the reparameterization trick and express the random vector $\rvz = q_\phi(\velementbefore, \velementafter)$ as a deterministic variable $\vz = g_\phi(\velementbefore, \velementafter, \rve)$, where $\rve$ is an auxiliary variable with independent marginal $p(\rve)$, and $g_\phi$ is a function parameterized by a neural net with parameters $\phi$. Details about how this function is specified are provided in Section \ref{sec:variational_neural_inferer}.

We can now rewrite the expectation term such that we can utilize a sampling-based method to estimate it. In addition to this, we set the prior distribution to be a Gaussian distribution $p(\rvz) \sim \mathcal{N}(\bm 0, \mI_n)$ and make our approximate posterior distribution $q(\rvz)$ also a normal distribution $\mathcal{N}(\vmu,\mSigma)$ with $\mSigma$ a diagonal matrix. Since the Kullback-Leibler divergence for these distributions has a closed form, we can write the following loss function:
\begin{align}
    \mathcal{L}(\theta,\phi,\velementafter)
     & = \frac{1}{2} \sum_{k=1}^d \left(1 + \log{(\sigma_k^2)} - \mu_k^2 - \sigma_k^2 \right) \nonumber         \\
     & \quad + \frac{1}{L} \sum_{l=1}^L \log{p_\theta(\velementafter |\velementbefore, \rvz^l)} \label{eq:loss}
\end{align}
In Equation \ref{eq:loss}, $L$ is the number of samples to take to obtain a good estimator of the expectation term. In principle, we follow previous work \cite{kingmaAutoEncodingVariationalBayes2014} and set the number of samples to 1 given that we train our model with a minibatch size that is large enough.

\subsection{Variational Neural Edit Encoder}

Edits are represented as sequences of tokens, such that $\elementbefore = [ x^{(1)}_-, \ldots, x^{(T)}_- ]$ and $\elementafter = [ x^{(1)}_+, \ldots,x^{(N)}_+]$. To obtain an edit representation, we further process these sequences using matching techniques \citep{yin_learning_2019} to obtain tags which identify the tokens that have been added ($+$), removed ($-$), replaced ($\Leftrightarrow$), or remained the same ($=$). In this process, as shown in Figure \ref{table:yin_matching}, we obtain padded versions of the sequences $\nelementbefore = [ \nelementbefore^{(1)}, \ldots, \nelementbefore^{(M)} ]$ and $\nelementafter = [ \nelementafter^{(1)}, \ldots, \nelementafter^{(M)} ]$, alongside with a sequence of tags $\tilde{x}_{tags}$ with length $M$ indicating the edit operations applied to each position. We denote the vocabulary for these tags as $\sV^{l} = \{-, +, =, \Leftrightarrow \}$ and the vocabularies for the tokens in $\nelementbefore$ and $\nelementafter$ as $\sV^{-}$ and $\sV^{+}$ respectively.

\begin{figure*}[h!]
    \scriptsize
    \centering
    \begin{tabular}{c c c c c c c c c c c c c}
        $\tilde{x}_-$:      & Disposal & of  & Waste             & material & according & to  & the & local & policies & ,                 & respectively & .      \\
        $\tilde{x}_+$:      & Disposal & of  & waste             & material & according & to  & the & local & policies & .                 & $\phi$       & $\phi$ \\
        $\tilde{x}_{tags}$: & $=$      & $=$ & $\Leftrightarrow$ & $=$      & $=$       & $=$ & $=$ & $=$   & $=$      & $\Leftrightarrow$ & $-$          & $-$    \\
    \end{tabular}
    \caption{Example of the edit matching pre-processing step. The example in this figure is taken from the QT21 De-En MQM dataset (index B1\_A6\_4w\_620). Its labels indicate that this post edit solves with problems related to spelling, typography, and the deletion of extra terms that are not needed.}
    \label{table:yin_matching}
\end{figure*}

We then separately embed the three sequences returned from the matching operation and perform element-wise concatenation to get $\tilde{\ve}$. We then feed $\tilde{\ve}$ to a bidirectional LSTM \citep{gravesFramewisePhonemeClassification2005, gravesSpeechRecognitionDeep2013} as follows:
\begin{align}
    \tilde{\ve}_i     & =
    \begin{bmatrix}
        \mE_+(\nelementbefore^{(i)})       \\
        \mE_-(\nelementbefore^{(i)})       \\
        \mE_{tags}(\tilde{x}_{tags}^{(i)}) \\
    \end{bmatrix}                                            \\
    \vec{\vh}_e^{(i)} & = \text{LSTM}(\vec{\vh}_e^{(i-1)}, \tilde{\ve}_i) \\
    \cev{\vh}_e^{(i)} & = \text{LSTM}(\cev{\vh}_e^{(i+1)}, \tilde{\ve}_i) \\
    \vh_e^{(i)}       & = [\vec{\vh}_e^{(i)}; \cev{\vh}_e^{(M-i)}]
\end{align}
In the equations above $\mE_+$, $\mE_-$, $\mE_{tags}$ are embedding matrices for $\nelementbefore$, $\nelementafter$, and $\tilde{x}_{tags}$, respectively. The bi-directional LSTM returns a sequence of hidden states or annotations. Each one of these can be seen as a contextualized, position-aware representation of the edit. We choose the last hidden state, $\vh_e^{(M)}$, as a fixed-length representation for the whole edit.

\subsection{Document Encoder}

To generate a fixed-length representation for each original document $\elementbefore$, we use another bidirectional LSTM as follows:
\begin{equation}
    \vh_d^{(i)} = \text{BiLSTM}(\vh_d^{({i-1})},\mE_-(\elementbefore^{(i)}))
\end{equation}
In a similar fashion to the variational edit encoder, we take the last hidden state as a fixed-length representation of $\elementbefore$.

\subsection{Variational Neural Inferer}
\label{sec:variational_neural_inferer}

As mentioned earlier, the posterior distribution is set to be a multivariate Gaussian parameterized by the mean and variance matrices. Specifically, we treat these as functions of both the original document $\elementbefore$ and the edited document $\elementafter$ as follows:
\begin{equation}
    g_\phi(\rvz | \velementbefore, \velementafter) \sim \mathcal{N}(\vmu_\phi(\elementbefore, \elementafter), \mSigma_\phi(\elementbefore, \elementafter))
\end{equation}
To approximate this posterior, we project the representation of the edit onto the latent space by using a linear projection layer to derive the vector $\vmu$ for the mean and another linear projection layer to derive a vector $\bf\sigma$ for the variance (we assume that $\mSigma_\phi$ is a diagonal matrix so we only need to estimate the values in its diagonal). We do this as follows:
\begin{align}
    \vmu               & = \mW_\mu \vh_e^{(M)}+ \vb_\mu        \\
    \log \bm{\sigma}^2 & = \mW_\sigma \vh_e^{(M)} + \vb_\sigma
\end{align}
In the equations above, $\mW_\mu \in \sR^{d_z \times d_e}$, $\mW_\sigma \in \sR^{d_z \times d_e}$ represent trainable weight matrices, and $\vb_\mu \in \sR^{d_z}$, $\vb_\sigma \in \sR^{d_z}$ represent the bias vectors of the linear projections we use. Finally, we can write the following:
\begin{equation}
    \vz = g_\phi(\velementbefore, \velementafter, \rve) := \vmu + \bm{\sigma} \odot \rve
\end{equation}
Here, $\rve \sim \mathcal{N}(0,\mI)$ is our introduced independent auxiliary random variable, and parameters of $g_\phi$ are therefore characterized by matrices $\mW_\mu$, $\mW_\sigma$ and bias vectors $\vb_\mu$ and $\vb_\sigma$.

For use during generation, we project our latent variable, $z$, to the target space with a linear transformation. We refer to this projected vector as $h'_e$. This is shown below:
\begin{equation}
    h'_e = W_e z + b_e
\end{equation}

\subsection{Variational Neural Editor}

To reconstruct $\elementafter$, we use a decoder which acts as a neural editor. This is implemented using another LSTM. This neural editor is conditioned both on the input document $\elementbefore$ and the edit representation $\rvz$, and it uses this information to apply the edit by generating $\elementafter$.


The procedure works as follows: (1) Firstly, the decoder is initialized with the concatenation of the projected latent vector and the representation of the original document $[\vz; \vh_d^{(T)}]$, (2) Since we want the decoder to reuse information from $\elementbefore$ as much as possible, the decoder attends its representation, making use of the set of annotation vectors $\vh_d$ on each timestep, (3) At each timestep, $\vh'_e$ is concatenated with the hidden state returned during the previous timestep as follows:
\begin{equation}
    \vh_d^{'(j)} = \text{LSTM}(\vh_d^{'(j-1)}, [\mE_+(\nelementafter^{(j)}); \vh'_e], \vc_j)
\end{equation}
The decoder's hidden state at timestep $j$ is referred as $\vh'_d{(j)}$ and the context vector $\vc_j = \sum_i \alpha_{ji} \vh_d$ is computed using general attention.

\subsection{The $x_{\Delta}$ loss}

Variational auto-encoders are often found to ignore latent variables when using flexible generators like LSTMs. Thus, in order to increase the likelihood of the latent space to be useful, we propose to encourage the latent vector to contain information about the tokens that have been changed (added, replaced, or removed), which we denote as $x_{\Delta}$.

Specifically, we require a decoder network to predict the set of tokens that have been changed in an unordered fashion. If we let $f =\text{MLP}(\vz) \in \mathcal{R}^{|\sV_+|}$, we have:
\begin{equation}
    \log p(x_{\Delta}|\rvz) = \log \prod_{t=1}^{|x_{\Delta}|}\frac{\exp{f_{x_t}}}{\sum_j^V \exp{f_j}}
\end{equation}
This term is added to Equation \ref{eq:loss}, and our model is trained to jointly minimize $-\log p(x_{\Delta}|\rvz)$ together with the rest of the loss terms.

\section{Experimental Setup}

\subsection{PEER: Performance Evaluation of Edit Representations}

Previous research on evaluating the quality of edit representations has mainly been by proposed \citet{yin_learning_2019}. We mainly find two kinds: \textit{intrinsic evaluations} of edit representations, for which no additional labels are required, and \textit{extrinsic evaluations}, which require additional labels or information.

A detailed revision of the existing literature in terms of \textit{intrinsic evaluations} showed us that this is performed mainly by measuring the gold-standard performance of the neural editor in terms of the average token-level accuracy, by visually inspecting the semantic similarity of neighbors in the latent space using human judgement, or by performing clustering and later visually inspecting some of the clusters obtained. We provide more details about each one of these techniques in our supplementary material. As can be seen, intrinsic evaluations are largely dependent on human studies, which are expensive and difficult to replicate. Instead of relying on this kind of evaluation, in this paper, we resort to automatic and more standard ways to do so. In addition to standard metrics used for generative models such as the cross entropy and BLEU-4, we propose the GLEU \cite{napolesGroundTruthGrammatical2015} evaluation metric. This metric was developed for the GEC task and is essentially is a variant of BLEU modified to account for both the source and the reference, making it more adequate for our task. It can also be interpreted as a more general version of the token-level accuracy metric utilized by \citet{yin_learning_2019}.

Regarding \textit{intrinsic evaluations}, we found that literature also offers a broad variety of alternatives. Among these, the most relevant included (1) visual inspection of the 2D-projected edit space, generally performed for a subset of the edits known to be associated to a certain label, (2) one-shot performance of the neural editor on similar edits ---previously identified by means of additional information---, and (3) the ability to capture other properties of the edit \cite{marrese-taylor_edit-centric_2019,sarkarStRESelfAttentive2019}, namely one or many labels associated to it.

Based on these findings, we propose a combination of training and evaluation datasets, each associated to a specific task in natural language processing, to automatically evaluate the quality of edit representations. We define a set of downstream tasks based on three different sources of edits, which we call \textbf{PEER} (Performance Evaluation for Edit Representations). Table \ref{table:datasets} provides a descriptive summary of the datasets included in PEER, and we provide details about each below:

\begin{table}[h!]
    \centering
    \scriptsize
    \begin{tabular}{c c c c c c}
        \toprule
        \textbf{Dataset} & \textbf{Size} & \textbf{Only $+$} & \textbf{Only $-$} & \textbf{Only $\Leftrightarrow$} & \textbf{Length} \\
        \midrule
        WikiAtomicSample & 104,000       & 50.0\%            & 50.0\%            & 0.0\%                           & 25.1            \\ 
        \midrule
        WikiEditsMix     & 113,983       & 24.1\%            & 16.6\%            & 34.0\%                          & 61.6            \\ 
        \midrule
        QT21 En-De       & 24,877        & 8.4\%             & 8.0\%             & 36.6\%                          & 20.0            \\ 
        \midrule
        QT21 En-De MQM   & 1,255         & 10.3\%            & 11.3\%            & 40.1\%                          & 19.2            \\ 
        \midrule
        Lang 8           & 498,359       & 13.2\%            & 4.6\%             & 45.9\%                          & 13.5            \\ 
        \midrule
        WI + Locness     & 25,556        & 11.9\%            & 4.1\%             & 42.3\%                          & 21.4            \\ 
        \bottomrule
    \end{tabular}
    \caption{Description of the datasets we utilize to train and evaluate our models.}
    \label{table:datasets}
\end{table}

\textbf{Wikipedia Edits}: We work with two large resources of human edits on Wikipedia articles.
\begin{itemize}
    \item \textbf{WikiAtomicSample}: We randomly sampled approximately $150K$ insertion and deletion examples from the English portion of the WikiAtomicEdits \citep{faruqui_wikiatomicedits:_2018}. After cleaning, we keep $104K$ samples.

    \item \textbf{WikiEditsMix}: We randomly selected 20 of the 200 most edited Wikipedia articles and extract the diff for each revision for each article using the WikiMedia API. We make use of the Wikimedia's ORES \citep{halfaker2019ores} API and scrape the \textit{draftquality} label for each revision. There are 4 \textit{draftquality} labels: \textit{spam, vandalism, attack}, and \textit{OK}, each corresponding to a different quality of the edit.
\end{itemize}
For this task, we evaluate the quality of the edit representations by means of running a multi-class classifier over the edit representations to predict the quality labels in the WikiEditsMix datasets. We use both datasets to train models.

\textbf{Post-editing}: As explained earlier, post-editing is the process whereby humans amend machine-generated translation. We choose one of the largest resources of human-annotated examples to train and evaluate our models.
\begin{itemize}
    \item \textbf{QT21 De-En}: We work with the German-English portion of the QT21 dataset \cite{speciaTranslationQualityProductivity2017}, which originally contains a total of 43,000 examples of machine translation human post-edits. The machine translation output over which post-editing is performed to create this dataset is an implementation of the attentional encoder-decoder architecture and uses byte-pair encoding \cite{sennrich_neural_2016}.
    \item \textbf{QT21 De-En MQM}: A subset of 1,800 examples of the De-En QT21 dataset, annotated with details about the edits performed, namely the reason why each edit was applied. Since the dataset contains a large number of edit labels, we select the classes that are present in at least 100 examples and generate a modified version of the dataset for our purposes. Examples where no post-edit has been preformed are also ignored.
\end{itemize}
The evaluation scheme on the post-editing task is based on the unlabeled data in QT21 De-En for training and the labeled data in the QT21 De-En MQM dataset for testing. Since each test example is associated to a variable number of labels, this task is cast as multi-label classification.

\textbf{Grammatical Error Correction (GEC)}: We consider the task of English GEC, which has attracted a lot of interest from the research community in the last few years. Since grammatical errors consist of many different types we follow previous work by \citet{bryantBEA2019SharedTask2019} and use some of the datasets released for this shared task, which work with well-defined subsets of error types.
\begin{itemize}
    \item \textbf{Lang-8 Corpus of Learner English (Lang 8)}: A corpus for GEC derived from the English subset of the Lang-8 platform, an online language learning website that encourages users to correct each other's grammar \cite{mizumotoEffectLearnerCorpus2012}. In particular, we work with the version of the dataset released by \citet{bryantBEA2019SharedTask2019} and further process it to skip examples where there are no grammar corrections.

    \item \textbf{W\&I + LOCNESS (WI + Locness)}: A dataset which was compiled by \citet{bryantBEA2019SharedTask2019}, built on top of (1) a subset of the LOCNESS corpus \cite{granger1998computer}, which consists of essays written by native English students manually annotated with grammar errors, and (2) manually annotated examples from the Write \& Improve online web platform \cite{yannakoudakisDevelopingAutomatedWriting2018}. This dataset contains 3,600 annotated examples across three different CEFR levels \cite{little_2006}: A (beginner), B (intermediate), and C (advanced). Again, we ignore examples where there are no grammar corrections.
\end{itemize}
The evaluation scheme for GEC consists on training models on the unlabeled Lang 8 dataset, and the evaluation is performed using the labels in WI + Locness, which associates CEFR difficulty levels to each example. Concretely, the problem is a multi-class classification problem.

\subsection{Comparison to prior work}

Using PEER as a test bed, we compare the performance of \textsc{EVE} against two relevant baselines. Firstly, we consider a variation of the deterministic encoder by \citet{yin_learning_2019}, with the only difference being that we do not include the copy mechanism in order to make results directly comparable.

\newcommand{\vmf}{\text{vMF}}
Secondly, we consider the approach by \citet{guuGeneratingSentencesEditing2018}. To adapt this to our setting, we skip the sampling procedure required in their case since our edits are already pairs of sentences and proceed to directly incorporate their edit encoding mechanism into our model. Following their approach, we first identify the tokens that have been added and removed for each edit, which we denote as $x_{\Delta}^+$ and $x_{\Delta}^-$. Each one of these token sequences is treated like a bag-of-words, encoded using trainable embedding matrix $\mE$, aggregated using sum pooling and finally projected using two different linear layers to obtain $\vh_+$ and $\vh_-$. These are finally combined to obtain $\vf = [\vh_+;\vh_-]$. In their approach, a sample from the approximate posterior $q$ is simply a perturbed version of $\vf$ obtained by adding von-Mises Fisher (vMF) noise, so they perturb the magnitude of $\vf$ by adding uniform noise.
\begin{align}
    q(\rvz_{\text{dir}} | x_{\Delta}^+, x_{\Delta}^-)  & = \vmf\left(\rvz_{\text{dir}} ; \vf_{\text{dir}}, \kappa\right)                                     \\
    q(\rvz_{\text{norm}} | x_{\Delta}^+, x_{\Delta}^-) & = \text{Unif}(\rvz_{\text{norm}} ; [\tilde{\vf}_{\text{norm}}, \tilde{\vf}_{\text{norm}}+\epsilon])
\end{align}
In the equations above, $\vf_\text{norm} = \|f\|$, $\vf_\text{dir} = \vf / \vf_\text{norm}$, $\vmf\left(v ; \mu, \kappa \right)$ denotes a vMF distribution with mean vector $\mu$ and concentration parameter $\kappa$, and $\tilde{\vf}_{\text{norm}} = \min(\vf_{\text{norm}}, 10 - \epsilon)$ is the truncated norm. Finally, the resulting edit vector is $\rvz = \rvz_\text{dir} \cdot \rvz_\text{norm}$, resulting in a model whose KL divergence does not depend on model parameters. We adapt their code release\footnote{\url{https://github.com/kelvinguu/neural-editor}} and integrate it into our codebase utilizing the same hyper-parameters. In order to make results comparable, we do not use pre-trained embeddings. For additional details, please refer to their paper and/or implementation.

\subsection{Implementation Details}

Despite the VAE's appeal as a tool to learn unsupervised representations through the use of latent variables, there exists the risk of ``posterior collapse'' \cite{bowmanGeneratingSentencesContinuous2016}. This occurs when the training procedure falls into the trivial local optimum of the ELBO objective, in which both the variational posterior and true model posterior collapse to the prior. This often means these models end up ignoring the latent variables, which is undesirable because an important goal of VAEs is to learn meaningful latent features for inputs. To deal with these issues, we utilize word dropout, and we anneal the KL term in the loss utilizing a sigmoid function, following the work of \citet{bowmanGeneratingSentencesContinuous2016}. Additionally, we also follow recent work of \citet{liSurprisinglyEffectiveFix2019}, who discovered that when the inference network of a text VAE is initialized with the parameters of an encoder that is pre-trained using an auto-encoder objective, the VAE model does not suffer from the posterior collapse problem. Therefore, our model is first trained with zero KL weight until convergence. Then, the decoder is reset, and the whole model is re-trained.

For the intrinsic evaluation, BLEU and GLEU scores are computed over the beam-search-generated output. Following the scheme of PEER, we first pre-train each model using the training scheme (i.e. the intrinsic task) and obtain a function that maps edits to a fixed-length vector representing a point in the latent space. We then evaluate this mapping function using the extrinsic evaluation setup. Since \textsc{EVE} is a probabilistic model, we utilize MAP and select the vector that parameterizes the mean of the posterior distribution as a deterministic edit representation for each example.

\section{Results}

To assess the contribution of each of our proposals, we performed an ablation study in two settings, WikiAtomicSample $\rightarrow$ WikiEditsMix and Lang 8 $\rightarrow$ WI + Locness. Specifically, we were interested in studying the effect on performance of our $x_{\Delta}$ loss, of the Kullback-Leibler divergence and of the pre-training technique proposed by \citet{liSurprisinglyEffectiveFix2019}. We evaluated each model variation in both the intrinsic and extrinsic tasks using the validation set on each case.

\begin{table}[h!]
    \centering
    \footnotesize
    \begin{tabular}{ccccc}
        \toprule
        \textbf{Data} & \textbf{Model}      & \textbf{BLEU} & \textbf{GLEU} & \textbf{Acc} \\
        \midrule
        \multirow{4}{*}{\shortstack{WikiAtomicSample                                       \\ $\rightarrow$ WikiEditsMix}}
                      & Base                & 0.81          & 0.79          & 0.672        \\
                      & + $x_{\Delta}$ loss & 0.82          & 0.80          & 0.767        \\
                      & + KL loss           & 0.77          & 0.75          & 0.649        \\
                      & EVE                 & \bf0.84       & \bf0.82       & \bf0.780     \\
        \midrule
        \multirow{4}{*}{\shortstack{Lang 8 $\rightarrow$                                   \\ WI + Locness}}
                      & Base                & 0.65          & 0.58          & 0.831        \\
                      & + $x_{\Delta}$ loss & 0.65          & 0.57          & 0.939        \\
                      & + KL loss           & 0.56          & 0.46          & 0.409        \\
                      & EVE                 & \bf0.68       & \bf0.61       & \bf0.958     \\
        \bottomrule
    \end{tabular}
    \caption{Results of our ablation studies. BLEU and GLEU scores are computed over the validation split, and Acc stands for accuracy of the respective downstream classification task for each dataset, also computed on the validation split.}
    \label{table:ablations}
\end{table}

Table \ref{table:ablations} summarizes the results of our ablation experiments. Results show the effectiveness of the introduced $x_{\Delta}$ loss, which consistently helps the baseline model obtain better performance. The fact that performance not only improves on the intrinsic tasks, but also on the extrinsic evaluation, suggests that this technique effectively helps the latent code store meaningful information about the edit. On the other hand, we see that the addition of the KL term to the loss tends to have negative effects on both the intrinsic and extrinsic tasks, evidencing the instability added by this constraint to the encoder. This result is not surprising, being consistent with previous findings in the context of text VAEs \cite{bahuleyanVariationalAttentionSequencetoSequence2018,liSurprisinglyEffectiveFix2019}. Finally, results of our full model show that both the $x_{\Delta}$ loss as well as the pre-training trick can be effectively combined to help the encoder stabilize and encourage the latent space to contain relevant information about the edits, leading to better performance overall.
\begin{table*}[h!]
    \centering
    \scriptsize
    \begin{tabular}{@{}cccccccccc@{}}
        \toprule
        \multirow{3}{*}{\textbf{Train. Data}} & \multirow{3}{*}{\textbf{Model}} & \multicolumn{4}{c}{\textbf{Intrinsic Evaluation}} &
        \multicolumn{4}{c}{\textbf{Extrinsic Evaluation}}                                                                                                                                                                                                                                                                                  \\
        \cmidrule{3-10}
                                              &                                 & \multicolumn{2}{c}{\textbf{Valid}}                & \multicolumn{2}{c}{\textbf{Test}} & \multirow{2}{*}{\textbf{Eval. Data}} & \multicolumn{3}{c}{\textbf{Accuracy}}                                                                                     \\
        \cmidrule{3-6} \cmidrule{8-10}
                                              &                                 & \textbf{BLEU}                                     & \textbf{GLEU}                     & \textbf{BLEU}                        & \textbf{GLEU}                         &                                 & \textbf{Train} & \textbf{Valid} & \textbf{Test} \\
        \midrule
        \multirow{3}{*}{WikiAtomicSample}     & Guu                             & 0.63                                              & 0.60                              & 0.28                                 & 0.26                                  & \multirow{6}{*}{WikiEditsMix}   & 0.738          & 0.740          & 0.743         \\
                                              & Yin                             & 0.81                                              & 0.79                              & 0.81                                 & 0.79                                  &                                 & 0.671          & 0.672          & 0.668         \\
                                              & EVE                             & \bf0.84                                           & \bf0.82                           & \bf0.84                              & \bf0.82                               &                                 & \bf0.782       & \bf0.780       & \bf0.774      \\
        \cmidrule{1-6} \cmidrule{8-10}
        \multirow{3}{*}{WikiEditsMix}         & Guu                             & 0.56                                              & 0.53                              & 0.54                                 & 0.52                                  &                                 & \bf0.670       & \bf0.668       & \bf0.666      \\
                                              & Yin                             & \bf0.65                                           & \bf0.65                           & \bf0.65                              & \bf0.65                               &                                 & 0.604          & 0.597          & 0.600         \\
                                              & EVE                             & 0.58                                              & 0.61                              & 0.55                                 & 0.57                                  &                                 & 0.637          & 0.642          & 0.638         \\
        \midrule
        \multirow{3}{*}{Lang 8}               & Guu                             & 0.53                                              & 0.43                              & 0.51                                 & 0.41                                  & \multirow{3}{*}{WI + Locness}   & 0.924          & 0.856          & 0.856         \\
                                              & Yin                             & 0.65                                              & 0.58                              & 0.65                                 & 0.58                                  &                                 & 0.836          & 0.831          & 0.831         \\
                                              & EVE                             & \bf0.68                                           & \bf0.61                           & \bf0.68                              & \bf0.60                               &                                 & \bf0.971       & \bf0.958       & \bf0.958      \\
        \midrule
        \multirow{3}{*}{QT21 De-En}           & Guu                             & 0.47                                              & 0.37                              & 0.32                                 & 0.30                                  & \multirow{3}{*}{QT21 De-En MQM} & 0.925          & 0.896          & 0.933         \\
                                              & Yin                             & \bf0.57                                           & \bf0.49                           & \bf0.57                              & \bf0.49                               &                                 & 0.972          & 0.952          & 0.964         \\
                                              & EVE                             & 0.53                                              & 0.45                              & 0.54                                 & 0.46                                  &                                 & \bf0.999       & \bf0.992       & \bf0.992      \\
        \bottomrule
    \end{tabular}
    \caption{Result of the intrinsic and extrinsic evaluations on our datasets, as defined by the PEER framework.}
    \label{table:results}
\end{table*}

Table \ref{table:results} shows our obtained results and compares them to relevant prior work by means of \textsc{PEER}. If we focus on the intrinsic evaluations, we can see that our approach is able to provide better performance in two datasets, with the deterministic baseline by Yin performing better elsewhere. Since these metrics are highly concerned with the reconstructive capabilities of the neural editor, we think this evidence mostly suggests that the \citet{guuGeneratingSentencesEditing2018} neural edit encoder is less capable of storing relevant information from the edits in the latent vector, which is probably because it depends only on the tokens that were modified, in an unordered manner.
\begin{figure}[hb!]
    \centering
    \includegraphics[width=\columnwidth]{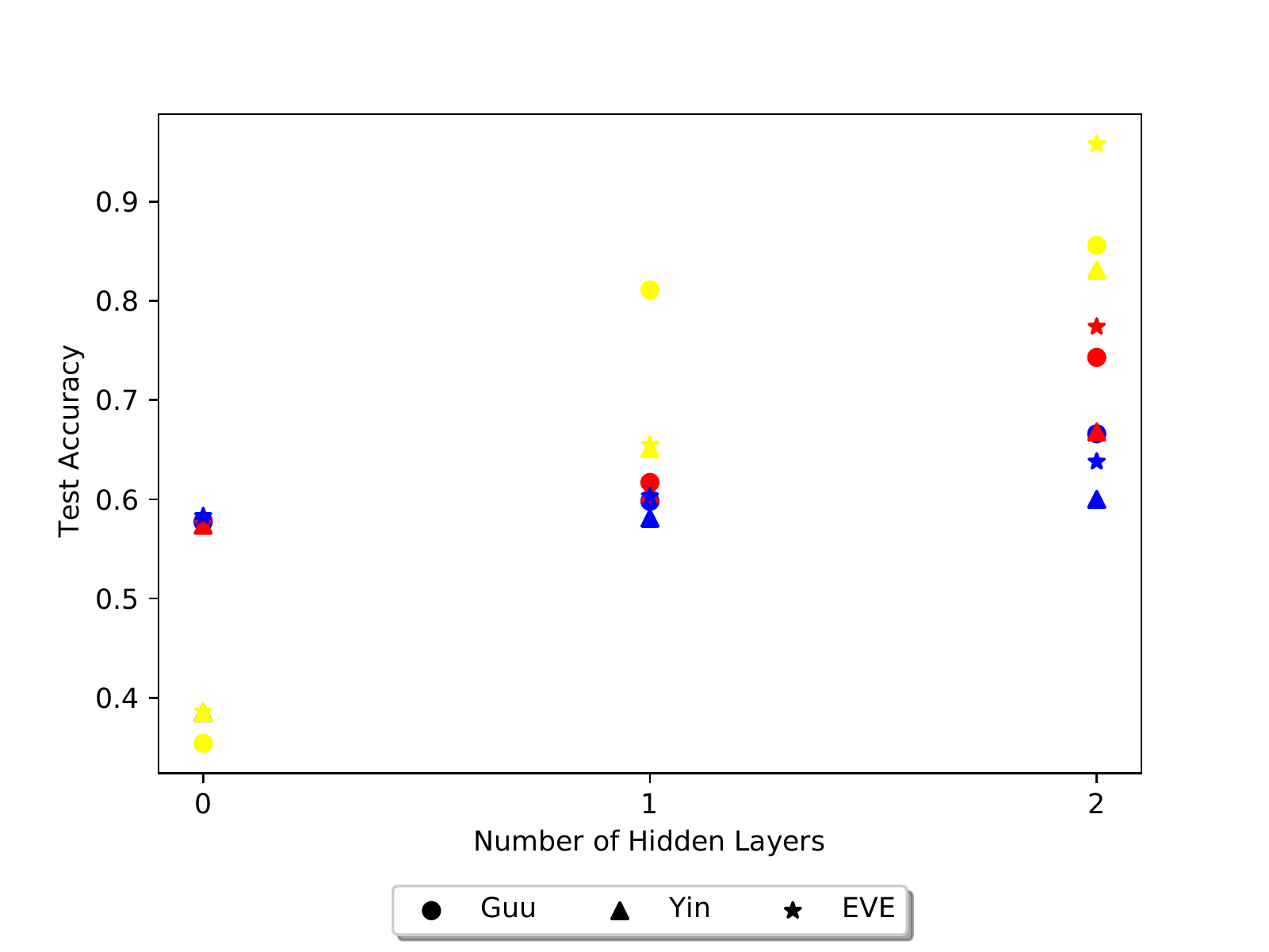}
    \caption{Effect of model depth on the accuracy on the test set of the extrinsic evaluation as a function of the number of hidden layers of the classifier. WikiAtomicSample $\rightarrow$ WikiEditsMix in red, WikiEditsMix$ \rightarrow$ WikiEditsMix in blue and Lang 8$\rightarrow$\ WI + Locness in yellow. Results on post editing are omitted for clarity.}
    \label{fig:peformance}
\end{figure}

Performance differences across settings are suggestive of the intrinsic difficulty of the task on each case. This is likely related to the nature of each dataset, which is evidenced in features such as average sentence length and vocabulary size. In this context, it is interesting to see models pre-trained on WikiAtomicSample outperforming models trained on WikiEditsMix. In this case, we think the much longer average sentence length may have hindered the learning process, since it is known that without the attention component, RNNs struggle on longer inputs \cite{bahdanau_neural_2015}.

In terms of the extrinsic evaluation, we see that our model obtains better performance in three out of the four settings, which we believe validates the effectiveness of our approach and shows that the information contained in our learned representations can actually be useful for downstream tasks. To select the best classifier, on each case we studied how performance varies with the depth of the classifier.

Finally, as Figure \ref{fig:peformance} shows, we see that at low depths, performance is poor, and differences across models tend to be small, suggesting that the classifiers are not capable of using the information stored in the vectors. Meanwhile, an increase in depth benefits all models, and as expected, it also allows us to clearly see the superiority of certain encoders. In this context, we think good results on WI + Locness and QT21 De-En MQM suggest that the labels in these are strongly correlated with the information stored in the edit representations, which is in agreement with the label nature (mostly related to the presence of certain misspelled/mistranslated terms). Conversely, the lower performance on WikiEditsMix suggests that a richer understanding of the edit semantics is needed.

\section{Conclusions}

In this paper, we have introduced a model that employs variational inference to learn a continuous latent space of vector representations to capture the underlying semantic information with regard to the document editing process. We have also introduced a set of downstream tasks specifically designed to evaluate the quality of edit representations, which we name PEER. We have utilized these to evaluate our model, compare it to relevant baselines, and offer empirical evidence supporting the effectiveness of our approach. We hope the development of PEER will help guide future research in this problem by providing a reliable programmatic way to test the quality of edit representations.


\section{Acknowledgements}
We are grateful to the NVIDIA Corporation, which donated two of the GPUs used for this research. We also thank Jorge Balazs, Pablo Loyola, Alfredo Solano, and Francis Zheng for their useful comments.

\bibliography{bibliography}

\end{document}